\documentclass{article}
\usepackage{spconf,amsmath,graphicx,hyperref}
\usepackage{multirow}
\usepackage{booktabs}
\usepackage{amssymb}
\usepackage{enumitem}

\makeatletter
\def\@maketitle{\newpage
 \null
 \vskip 2em \begin{center}
 {\large \bf \@title \par} \vskip 1.5em {\large \lineskip .5em
 \@name \@address
 \par} \end{center}
 \par
 \vskip 1.5em}
\makeatother

\title{A Comprehensive Ecosystem for\\Open-Domain Customized Video Generation}
\name{
Jingxu Zhang$^{1,2,*}$,
Yuqian Hong$^{1}$,
Daneul Kim$^{3}$,
Kai Qiu$^{2}$,
Qi Dai$^{2}$,  \\
\textit{Jianmin Bao}$^{2}$,
\textit{Yifan Yang}$^{2}$,
\textit{Xiaoyan Sun}$^{1,\dag}$,
\textit{Chong Luo}$^{2,\dag}$
}

\address{
$^{1}$University of Science and Technology of China\quad
$^{2}$Microsoft Research Asia\quad\\
$^{3}$Seoul National University\\
$^{*}$Work done during internship at Microsoft Research Asia.\quad
$^{\dag}$Corresponding authors.
}

\begin{document}
\ninept
\maketitle
\begin{abstract}
Recent progress in video generation has shown impressive visual synthesis capabilities.
However, open-domain customized video generation remains limited by the lack of large-scale, annotated datasets capturing diverse identity-specific attributes.
To address this, we introduce PexelsCustom-1M, the first publicly available million-scale dataset for identity-preserving video generation, containing one million curated $\langle identity, text, video \rangle$ triplets across 8,000+ categories.
Leveraging this, we propose CustoMDiT, a parameter-efficient framework that adapts a pretrained multimodal Diffusion Transformer into a customized video generator with only 8\% additional learnable parameters.
Our method surpasses prior state-of-the-art.
However, benchmarks such as DreamBooth cover only 100 classes, which is insufficient for real-world applications.
To overcome this, we construct OpenCustom, a new benchmark with 1,000+ categories, created via cross-dataset knowledge fusion from ImageNet and MS-COCO.
Extensive experiments confirm the advantages of both our dataset and model.
We will open-source the entire ecosystem—including dataset, pipeline, benchmark, and implementations—to support further research.
\end{abstract}
\section{Introduction}
\label{sec:intro}

The rapid advancement of video generation has intensified demands for customizable content creation in domains such as advertising and digital media. Customized Video Generation (CVG) seeks to preserve visual identities while embedding them into diverse scenarios guided by text. Although prior works in customized image \cite{dreambooth, gal2022image,ipadapter, msdiffusion, blipdiffusion, ominicontrol, ssrencoder, t2iadapter, controlnet, elite} and video generation \cite{motionbooth, dreamvideo, customcrafter, videobooth, idanimator, consisid, stillmoving, animatediff, dreamvideo2} have shown promise, they remain limited by (1) narrow categorical scope \cite{idanimator, consisid, videobooth, dreamvideo2} or (2) reliance on test-time optimization with per-identity fine-tuning \cite{dreambooth, motionbooth, dreamvideo, customcrafter}. These constraints stem from the lack of large-scale multimodal data linking diverse identities with contextual descriptions.

In this paper, we introduce PexelsCustom-1M, the first large-scale open-domain single-reference CVG dataset, curated via a dual-phase pipeline. Starting from 400K HD Pexels\footnote{https://www.pexels.com} videos, the preprocessing stage applies vision-language captioning, identity extraction, localization, and segmentation to establish precise identity-video correspondences. The postprocessing stage further refines samples through multi-stage filtering, subject-centric re-captioning with contextual preservation, and augmentation to mitigate artifacts. This workflow yields one million high-quality $\langle identity, text, video \rangle$ triplets across 8,000+ categories, enabling unprecedented scale and contextual diversity.

Built on PexelsCustom-1M, we propose CustoMDiT, a parameter-efficient Diffusion Transformer framework for CVG. CustoMDiT conditions text-to-video generation on identity-aware reference images via bias-injected RoPE embeddings, while LoRA layers enable efficient adaptation with minimal additional parameters. Experiments demonstrate superior performance over existing methods on standard CVG benchmarks, together with  improved efficiency.

Existing CVG benchmarks cover only \~100 categories, limiting generalization. 
To address this, we propose OpenCustom, a comprehensive evaluation suite spanning 1,000+ categories by fusing ImageNet-1K \cite{russakovsky2015imagenet} and MS-COCO \cite{lin2014microsoft}. OpenCustom provides a unified protocol for (1) identity extraction, (2) context-aware prompting, and (3) multi-dimensional evaluation.

Extensive experiments on both prior and our new benchmark demonstrate the superiority of PexelsCustom-1M and CustoMDiT. Human studies further confirm our results match or surpass competing approaches, including commercial CVG APIs (e.g., Vidu\footnote{https://www.vidu.cn/create/character2video}). We will open-source the dataset, curation pipeline, benchmark, and model to advance community research.
We summarize our key contributions:

\begin{itemize}
\item PexelsCustom-1M: The first large-scale, publicly available dataset for CVG, providing 1M $\langle identity, text, video \rangle$ triplets across 8,000+ identity categories.
\item Scalable Data Pipeline: A reproducible framework for harvesting and refining identity-text-video triplets, extensible to broader domains.
\item CustoMDiT: A parameter-efficient customized video generation model achieving state-of-the-art CVG performance with minimal architectural overhead.
\item OpenCustom Benchmark: A comprehensive evaluation protocol spanning 1,000+ categories to assess open-domain generalization in realistic settings.
\end{itemize}

\begin{figure*}
    \centering
    \includegraphics[width=1.0\linewidth]{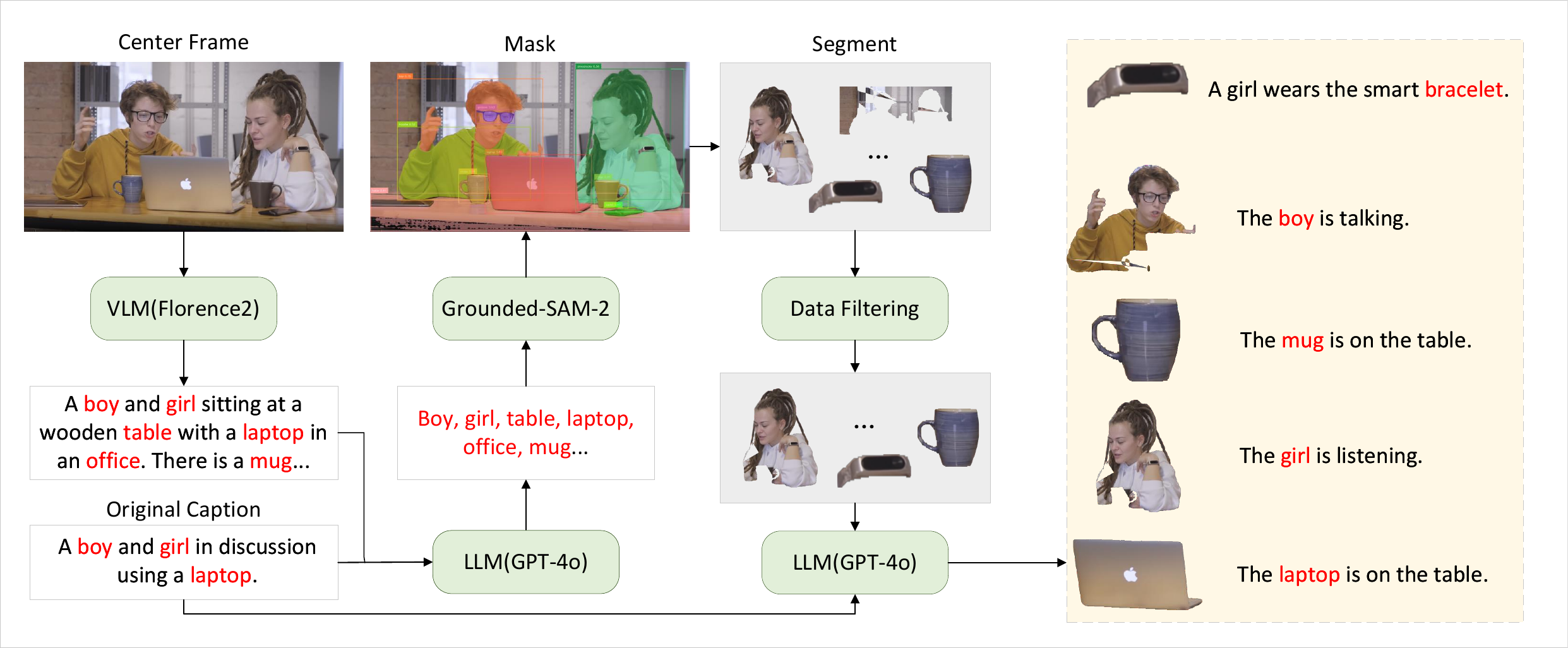}
    \vspace{-2mm}
    \caption{Data curation pipeline of our PexelsCustom-1M. We aim at enriching dataset during pre-processing, while filtering and generating subject-centric caption during post-processing.}
    \vspace{-3mm}
    \label{fig:data_pipe}
\end{figure*}

\section{Open-Domain Data Curation}
\label{sec:data}

\subsection{Data Pre-Processing}
\label{sec:preprocessing}
Pexels-400K contains high-quality videos, each accompanied by a descriptive caption. However, these captions primarily focus on the main subject and its motion, while lacking descriptions of other present identities. To address this limitation, we employ a vision-language model (VLM) \cite{xiao2023florence} to generate subject-centric captions for the center frame of each video. Since VLMs naturally emphasize identifying multiple entities in image captioning, we can extract additional identities from videos by designing appropriate prompts.

Next, we use GPT-4o to extract identities from both the original and subject-centric captions while filtering out background elements. Following previous mask generation methods \cite{videobooth, dreamvideo2}, we apply Grounded-SAM \cite{gsam} to generate masks for each extracted identity. Specifically, the extracted identities and the center frame of each video are first processed by Grounding-DINO \cite{ren2024grounding} to obtain bounding boxes, which are then refined by SAM \cite{kirillov2023segment} to produce segmentation masks. Our data curation pipeline is shown in Fig.~\ref{fig:data_pipe}. 

\subsection{Data Post-Processing}

\label{sec:postprocessing}

\textbf{Data Filtering.} To ensure the quality of reference images, we implement a series of carefully designed data filtering strategies, including aesthetic filtering, bbox size filtering, overlapped object filtering, etc.

\textbf{Re-Captioning.} The identities extracted by VLM during pre-processing lack corresponding captions. While we could simply use VLM-generated captions or append identities to the original caption, these approaches either lack motion descriptions or detailed identity context. Instead, as illustrated in Fig.~\ref{fig:data_pipe}. We input the identity name, original caption,center frame and cropped reference image to GPT-4o to generate a new caption that focuses on the identity while preserving the original caption's information.Caption–identity consistency is improved through re-captioning, raising the subject-identity CLIP score from 22.24 to 23.27.

\textbf{Data Augmentation.} 
The most effective approach to mitigating the copy-paste problem is cross-pair data training \cite{moviegen}. Additionally, we find that data augmentation during training further alleviates this issue. Specifically, we apply random resizing, rotation, and shifting to identities during training. Furthermore, we introduce a random shift in the frame sampling strategy to prevent the reference frame from being tied to a specific frame. The data augmentation effectively reduces the copy-paste problem.

\subsection{Data Statistics}
\label{sec:statistics}

\begin{table}[htbp]\small
  \centering
  \begin{tabular}{@{}lclclc@{}}
    \toprule
    Name & Open-Source & Type & $N_c$ & $N_s$ \\
    \midrule
    Dreambooth\cite{dreambooth} & \checkmark & Image & 30 & 158 \\
    CustomConcept\cite{customdiffusion} & \checkmark & Image & 101 & 634 \\
    Subjects200K\cite{ominicontrol} & \checkmark & Image & 4696 & 224K \\
    CustomStudio\cite{customvideo} & $\times$ & Image & 13 & 68 \\
    \midrule
    VideoBooth\cite{videobooth} & \checkmark & Video & 9 & 49K \\
    DreamVideo-2\cite{dreamvideo2} & $\times$ & Video & 2538 & 230K \\
    PexelsCustom (ours) & \checkmark & Video & 8373 & 1M \\
    \bottomrule
  \end{tabular}
  \caption{Statistics comparison on different Custom Image/Video Dataset. $N_c$ represents the number of classes, while $N_s$ represents the number of samples.}
  \label{tab:statistics}
\end{table}
We report a comparison of the statistics of our dataset with other existing image/video customization datasets in Table ~\ref{tab:statistics}. As the results show, the only open-source video customization dataset is from VideoBooth. However, the number of classes in the VideoBooth dataset is extremely limited to just 9, and the total number of samples is also insufficient. In contrast, we have collected our dataset with a significantly wider domain and larger scale, even compared to closed-source video customization datasets.

\section{Method}
\label{sec:method}

\begin{figure}
    \centering
    \includegraphics[width=1.0\linewidth]{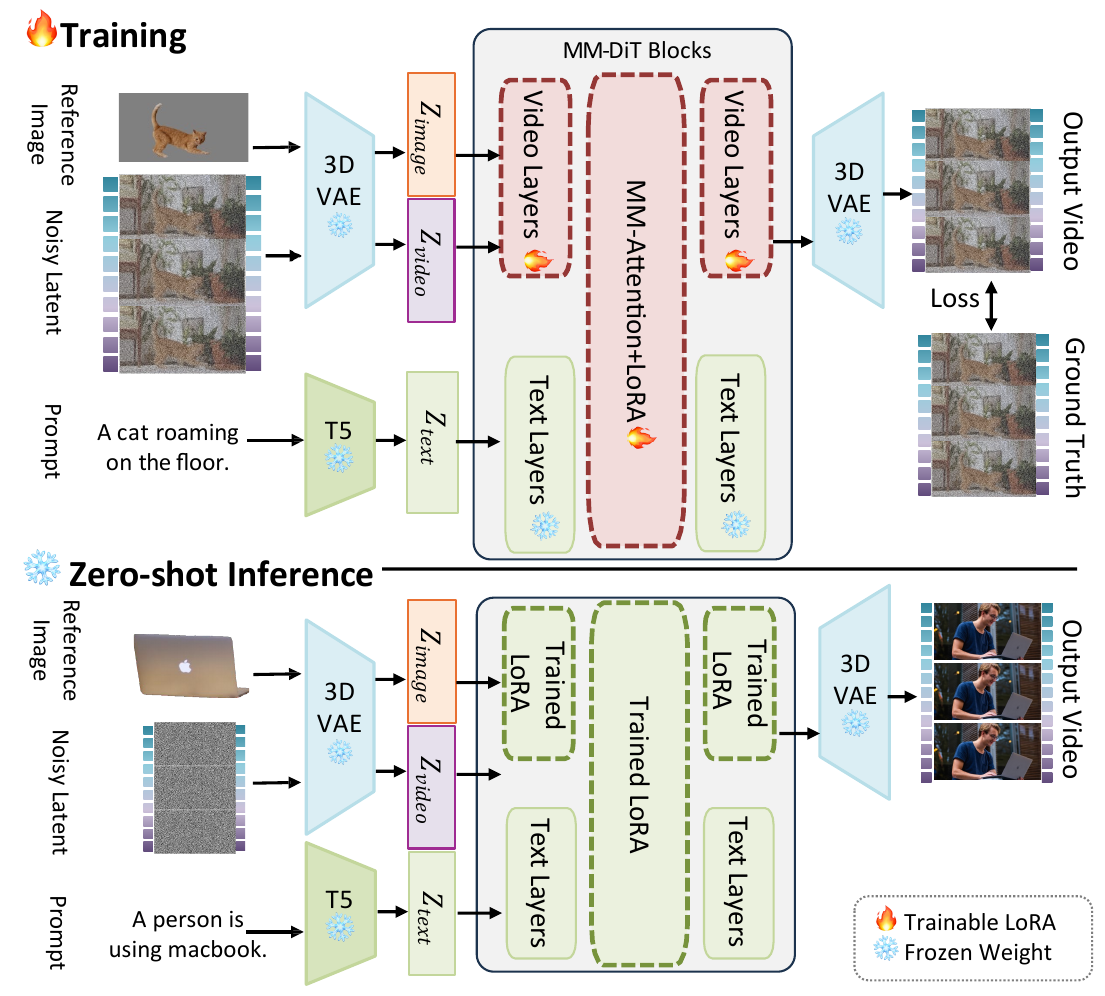}
    \vspace{-2mm}
    \caption{\textbf{Overview of CustoMDiT.} We demonstrate training pipeline (above) and how we conduct inference in zero-shot manner (below). The module enclosed in red dashed lines are equipped with LoRA layers and trained as shown in the figure above. Equipped with the trained LoRA, we conduct inference with the given reference image and prompt.}
    \label{fig:model}
    \vspace{-4mm}
\end{figure}

Fig.~\ref{fig:model} summarizes the training and inference pipeline of CustoMDiT. Following OminiControl~\cite{ominicontrol}, we inject the reference image via a Low-Rank Adapter (LoRA) while keeping the pretrained backbone frozen. Prior approaches~\cite{ipadapter, elite, videobooth} typically rely on a learned feature extractor or an off-the-shelf image encoder (e.g., CLIP), which often emphasizes high-level semantics and makes fine-detail injection difficult. Instead, we extract reference features using the pretrained 3DVAE. To keep the model subject-focused (rather than drifting toward an image-to-video behavior), we gray-pad the masked background of the reference image.

Given the modality fusion design of MM-DiT, we incorporate reference latents by reusing the video layers. We attach LoRA to all linear layers and attention projections in these video layers, process reference latents through the same layers, and concatenate them with video latents within attention. Importantly, LoRA is enabled only for reference-latent processing and disabled for video-latent processing, encouraging the adapters to specialize in reference feature injection.

\section{Experiments}
\label{sec:exp}

\begin{figure*}[h]
    \centering
    \includegraphics[width=1.0\linewidth]{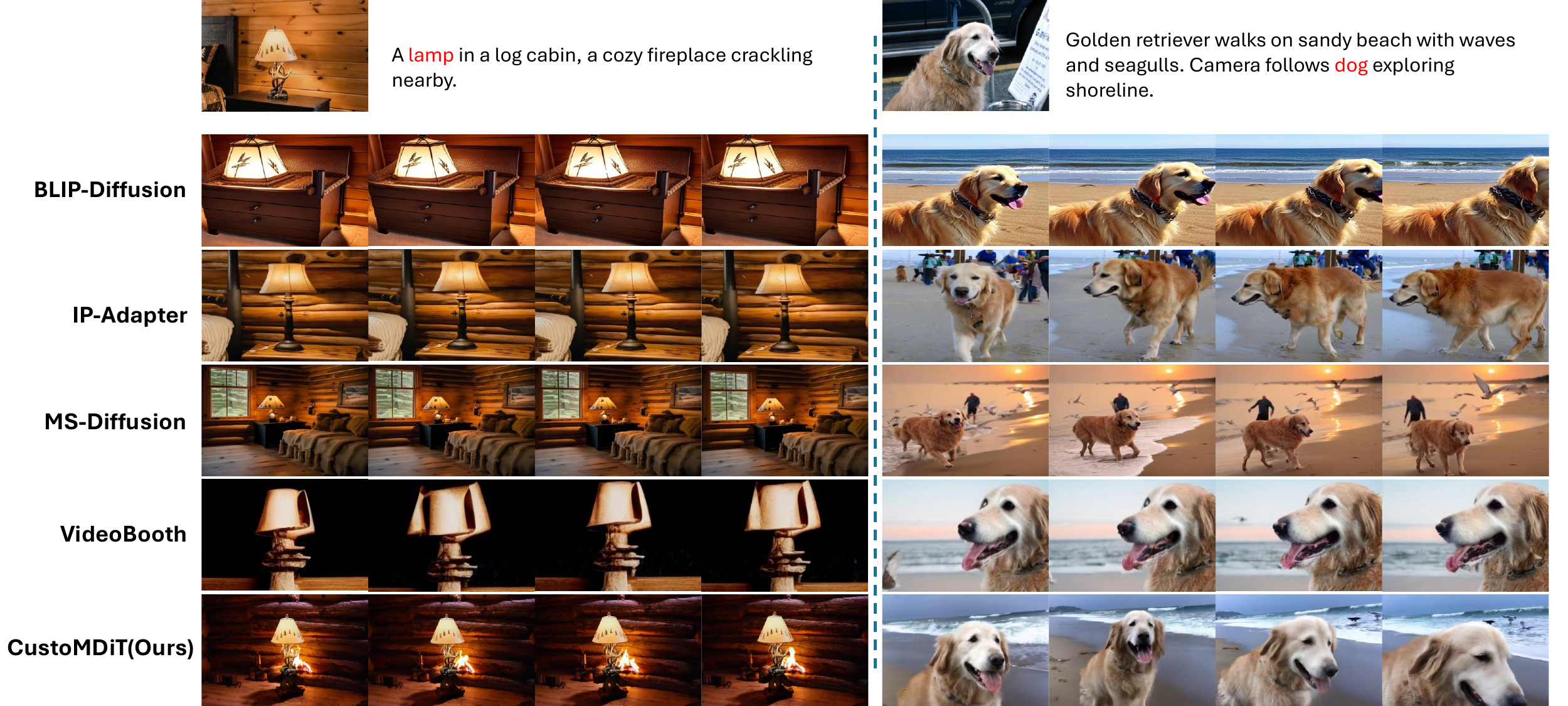}
    \caption{Qualitative Comparison with previous methods. VideoBooth is a PT2V method while the other three methods are implemented in a naive PT2I + I2V pipeline.}
    \label{fig:qualitative}
    \vspace{-4mm}
\end{figure*}

\begin{table}[t]\scriptsize
    \centering
    \setlength{\tabcolsep}{3pt}
    \resizebox{\columnwidth}{!}{
    \begin{tabular}{l|l|ccccc}
        \toprule
        \textbf{Method} & \textbf{Dataset} 
        & \textbf{M.S.} & \textbf{D.D.} & \textbf{CLIP-T} & \textbf{CLIP-I} & \textbf{DINO-I} \\
        \midrule

        \multirow{2}{*}{OminiControl\cite{ominicontrol}}
        & DreamBooth-Custom & 98.91 & 24.00 & \underline{30.45} & 68.64 & 43.69 \\
        & OpenCustom        & 98.69 & 38.64 & 31.34 & 64.57 & 34.69 \\
        \midrule

        \multirow{2}{*}{MS-Diffusion\cite{msdiffusion}}
        & DreamBooth-Custom & \underline{99.25} & 9.00 & 30.08 & 76.21 & \underline{62.67} \\
        & OpenCustom        & \underline{98.90} & 20.36 & \textbf{31.51} & 75.23 & \underline{59.44} \\
        \midrule

        \multirow{2}{*}{BLIP-Diffusion\cite{blipdiffusion}}
        & DreamBooth-Custom & 98.93 & 4.00 & 27.64 & 76.36 & 54.21 \\
        & OpenCustom        & 98.51 & 20.93 & 28.79 & \textbf{76.05} & 54.60 \\
        \midrule

        \multirow{2}{*}{IP-Adapter\cite{ipadapter}}
        & DreamBooth-Custom & 98.93 & 7.00 & 28.96 & \underline{76.52} & 54.57 \\
        & OpenCustom        & 98.62 & 29.50 & 30.86 & 74.21 & 49.00 \\
        \midrule

        \multirow{2}{*}{VideoBooth\cite{videobooth}}
        & DreamBooth-Custom & 96.97 & \underline{50.00} & 27.25 & 61.63 & 31.38 \\
        & OpenCustom        & 96.61 & \underline{57.86} & 28.20 & 67.69 & 41.48 \\
        \midrule

        \multirow{2}{*}{ID-Animator\cite{idanimator}}
        & DreamBooth-Custom & \textbf{99.30} & 5.00 & \textbf{30.94} & 67.29 & 34.62 \\
        & OpenCustom        & \textbf{99.14} & 8.79 & \underline{31.50} & 66.81 & 34.38 \\
        \midrule

        \multirow{2}{*}{CustoMDiT (ours)}
        & DreamBooth-Custom & 97.66 & \textbf{61.00} & 29.17 & \textbf{76.93} & \textbf{66.59} \\
        & OpenCustom        & 97.42 & \textbf{70.29} & 30.96 & \underline{75.32} & \textbf{65.80} \\

        \bottomrule
    \end{tabular}
    }
    \caption{Benchmark results on two datasets. The best method is in \textbf{bold}, and the second-best is \underline{underlined}.}
    \label{tab:benchmark_merged}
\end{table}

\subsection{Experimental Setup}
\label{sec:exp_setup}

\textbf{Implementation Details.} 
We use CogVideoX-5B \cite{cogvideox} as the base model for CustoMDiT, setting both the LoRA rank and LoRA alpha to 128. CustoMDiT is trained on PexelsCustom-1M for 8,000 steps (global batch size 128) without data augmentation, using 64 NVIDIA A100 GPUs for 60 hours, followed by an additional 2,000 training steps with data augmentation. We use resolution of \(480\times720\) with 49 frames at 8 FPS. The text drop rate is fixed at 0.1. 

We employ no learning rate scheduler and optimize using AdamW with a learning rate of \(1\times10^{-4}\), betas of \([0.9, 0.95]\), and epsilon of \(1\times10^{-8}\). Denoising is performed using DPM as the noise scheduler, with 50 denoising steps during inference. Additionally, we utilize text classifier-free guidance scale of 6.0.

\textbf{Comparison Methods.} 
We compare our method with zero-shot PT2V and PT2I approaches. For PT2I, we evaluate our model against broad-domain techniques, including OminiControl \cite{ominicontrol}, MS-Diffusion \cite{msdiffusion}, BLIP-Diffusion \cite{blipdiffusion}, and IP-Adapter \cite{ipadapter}. As a naive baseline for generating customized videos, we apply CogVideoX-5B-I2V \cite{cogvideox} to the generated customized images. For PT2V, we compare our approach with VideoBooth \cite{videobooth}.

\subsection{Evaluation Benchmark}
As there is no publicly available or widely recognized benchmark for video customization, we evaluate the models using our own curated datasets. To assess the open-set generation capability of the models, we construct an evaluation dataset based on various image datasets.

\textbf{DreamBooth-Custom}
Following prior works \cite{motionbooth, videobooth, dreamvideo2, customcrafter}, we select 30 subjects from the DreamBooth dataset \cite{dreambooth} and 70 from the CustomConcept101 dataset \cite{customdiffusion}, sampling one image per subject. Each concept’s prompt is randomly chosen from the provided examples.

\textbf{OpenCustom benchmark}
We build OpenCustom—an open-domain benchmark using ImageNet \cite{imagenet} and MS-COCO \cite{coco}. For ImageNet-1K, all 1,000 classes are used by manually selecting one high-resolution image per class with a prominent subject matching the class name; GPT-4o generates prompts (motion prompts for living creatures, camera motion for objects), forming ImageNet-Custom. For MS-COCO, five subjects per category (totaling 400 samples) are manually chosen—only those with a single, prominent instance and minimal interference are retained; GPT-4o similarly generates the prompts.

\textbf{Evaluation Metrics} 
We evaluate our method using seven metrics, covering two key aspects of quality assessment. For identity preservation, we apply CLIP Image Similarity (CLIP-I) \cite{clip} and DINO Image Similarity (DINO-I) \cite{dino}.For video quality and consistency, we introduce three additional metrics. CLIP Text Similarity (CLIP-T) is used to assess the model’s ability to follow prompts. Motion Smoothness (M.S.) and Dynamic Degree (D.D.), derived from VBench \cite{vbench}, evaluate motion consistency and modeling capability.

\subsection{Experiment Results}
\textbf{Qualitative Comparison}
As shown in Fig.~\ref{fig:qualitative}, our method excels in both identity preservation and dynamic motion. On the left, it uniquely retains fine details—such as the lamp’s shape and texture—and is the only one to generate a crackling fireplace as prompted. On the right, while VideoBooth preserves the dog’s identity similarly, it falls short in capturing motion dynamics and following the text prompt. Moreover, although PT2I + I2V methods deliver strong prompt adherence and high aesthetic quality in initial frames, they often neglect the reference image and struggle with action consistency and camera movement.

\textbf{Quantitative Comparison}
We present quantitative results in Table ~\ref{tab:benchmark_merged}. CustoMDiT achieves state-of-the-art identity preservation across all benchmarks, with improved motion dynamics and competitive motion smoothness. The gains in DINO-I scores indicate its strong capacity to capture fine-grained subject details, complementing CLIP’s emphasis on semantic similarity. ID-Animator focuses on human face customization and does not generalize well, showing extremely low dynamic degree. Our method also outperforms VideoBooth and matches PT2I approaches in CLIP-T, confirming superior prompt adherence in both background consistency and motion generation. In contrast, image customization methods adapted with I2V models exhibit poor motion dynamics despite text conditioning. Overall, these results validate the versatility of our approach in open-domain and real-world scenarios.

\textbf{Human evaluation}
We conducted a user study to compare our method with VideoBooth \cite{videobooth}, OminiControl + I2V \cite{ominicontrol}, and the commercial video customization model Vidu2.0. Each participant was presented with 20 groups of videos, where each group contained four videos generated by the different methods in a randomized order. Participants were asked to evaluate the videos based on four criteria: (1) ID consistency, (2) Prompt alignment, (3) Motion quality, and (4) Overall quality.
\begin{table}[h]\small
    \centering
    \begin{tabular}{l|cccc}
        \toprule
        \textbf{Method} & \textbf{Prompt} & \textbf{ID} & \textbf{Motion} & \textbf{Overall} \\
        \midrule
        VideoBooth\cite{videobooth}   & 1.913 & 2.103 & 1.888 & 1.550 \\
        OminiControl\cite{ominicontrol} & 3.232 & 2.805 & 3.540 & 2.983 \\
        Vidu2.0\cite{vidu}      & 3.792 & 4.062 & 3.878 & 3.688 \\
        CustoMDiT (ours)    & \textbf{3.973} & \textbf{4.445} & \textbf{3.895} & \textbf{3.833} \\
        \bottomrule
    \end{tabular}
    \caption{Result on human evaluation, our method achieves best score on all aspects.}
    \label{tab:user_study}
\end{table}

A total of 30 participants took part in the study, and the results, shown in Table ~\ref{tab:user_study}, indicate that our method achieved the highest scores across all four aspects. Notably, we significantly outperformed OminiControl and VideoBooth while also surpassing the commercial Vidu2.0 model.

\subsection{Ablation Studies}

\textbf{Subject-Centric Re-Captioning} 
To evaluate the effectiveness of our enriched data with subject-centric re-captioning, we conducted an ablation study by removing the newly extracted subject information from the captions generated by the vision-language model (VLM) and replacing them with the original captions. As shown in Table ~\ref{tab:abl_dataset}, our mixed dataset, which combines original and newly extracted subjects, achieves comparable results to the non-re-captioned dataset while enhancing motion dynamics.

\begin{table}[t]\small
  \centering
  \begin{tabular}{@{}l|ccccc@{}}
    \toprule
    \textbf{Setting} & \textbf{M.S.} & \textbf{D.D.} & \textbf{CLIP-T} & \textbf{CLIP-I} & \textbf{DINO-I} \\
    \midrule
    Default & 97.39 & 67.00 & 29.34 & 75.34 & 63.52 \\
    No recaption   & 97.90 & 48.00 & 29.51 & 75.56 & 64.19 \\
    \bottomrule
  \end{tabular}
  \caption{Results for recaption ablation study.}
  \label{tab:abl_dataset}
\end{table}

\section{Conclusion}
\label{sec:conclusion}

We present a large-scale open-domain dataset for customized video generation (CVG). Building on this, we develop an efficient CVG framework via LoRA-adapted MMDiT. To rigorously evaluate open-domain generalization, we introduce a benchmark covering over 1,000 categories. We will open-source all resources to support future research.
While our method advances CVG capabilities, there are some limitations: (1) Performance is inherited from the pretrained MMDiT model, and a stronger base model could lead to a better performance; (2) Current focus on single-identity generation leaves multi-identity scenarios unexplored; (3) Cross-paired data could further enhance compositionality beyond our current data augmentation strategies. We plan to explore them in the near future.

\bibliographystyle{IEEEbib}
\bibliography{refs_short}

\end{document}